\documentclass[letterpaper, 10 pt, conference]{IEEEtran}
\IEEEoverridecommandlockouts
\usepackage{booktabs}
\usepackage{adjustbox}
\usepackage{cite}
\usepackage[hyphens]{url}

\usepackage{amsmath,amssymb,amsfonts}
\usepackage{algorithm}
\usepackage{algpseudocode}
\usepackage{graphicx}
\usepackage{textcomp}
\usepackage{hyperref}
\usepackage{xcolor}
\usepackage{multirow}
\usepackage{microtype}

\usepackage{enumitem}
\usepackage[T1]{fontenc}

\newif\ifarxiv
\arxivtrue

\begin{document}

\title{Cost-Aware Kinematically Feasible Planning \\ for Mobile and Surface Robotics
\author{Steve Macenski$^1$, Matthew Booker$^2$, Joshua Wallace$^3$, and Tobias Fischer$^4$
\thanks{$^1$Open Navigation LLC (e-mail: {\tt\footnotesize steve@opennav.org}).}%
\thanks{$^2$Carnegie Mellon University (e-mail: {\tt\footnotesize mrbooker@andrew.cmu.edu}).}%
\thanks{$^3$Locus Robotics (e-mail: {\tt\footnotesize jwallace@locusrobotics.com}).}%
\thanks{$^4$(Corresponding author) QUT Centre for Robotics and School of Electrical Engineering and Robotics, Queensland University of Technology, Brisbane, Australia. (e-mail: {\tt\footnotesize tobias.fischer@qut.edu.au}).}%
\thanks{This research was partially supported by funding from ARC DECRA Fellowship DE240100149 to TF and the QUT Centre for Robotics. We would like to thank Locus Robotics for their generous assistance in providing the warehouse map, and RoboTech Vision for their detailed assistance.}%
}
}

\maketitle
\begin{abstract}
We present \textit{Smac Planner}, an openly available, search-based planning framework that addresses the critical need for kinematically feasible path planning across diverse robot platforms. %
\textit{Smac Planner} provides high-performance implementations of Cost-Aware A*, Hybrid-A*, and State Lattice planners that can be deployed for Ackermann, legged, and other large non-circular robots. Our framework introduces novel ``Cost-Aware'' variations that significantly improve performance in complex environments common to mobile robotics while maintaining kinematic feasibility constraints. Integrated as the standard planning system within the popular ROS 2 Navigation stack, Nav2, \textit{Smac Planner} now powers thousands of robots worldwide across academic research, commercial applications, and field deployments.%
\end{abstract}


\section{Introduction}
\label{sec:introduction}
\IEEEPARstart{C}{ommercially} deployed mobile robots often still rely on classical techniques such as Dijkstra's Algorithm~\cite{dijkstra1959note} and numerous forms of grid-based heuristic search such as A*~\cite{A} and D*~\cite{diffA*}. For circular differential-drive or holonomic robots, this class of algorithms is still suitable for many uses. However, these planners produce paths that are physically impossible to follow when applied to robots that cannot execute arbitrary motions, such as Ackermann-steered vehicles, legged and humanoid platforms, and large non-circular robots.

Seminal works~\cite{pivtoraiko2009differentially,lavalle2001randomized,sbpl,karaman2011sampling,howard2007optimal} have established theoretical foundations for kinematically feasible planning, but they lack quality implementations accessible for academic use, integration with major research frameworks (MRPT~\cite{mrpt}, ROS~\cite{ros,macenski2022robot}, etc.), and variants tailored to mobile robotics requirements. This results in a large gap in easily deploying robots in emerging applications like last-mile delivery, construction monitoring, agriculture, and marine systems where these types of platforms are increasingly common.

\begin{figure}[ht]
    \centering
    \includegraphics[width=0.8\linewidth]{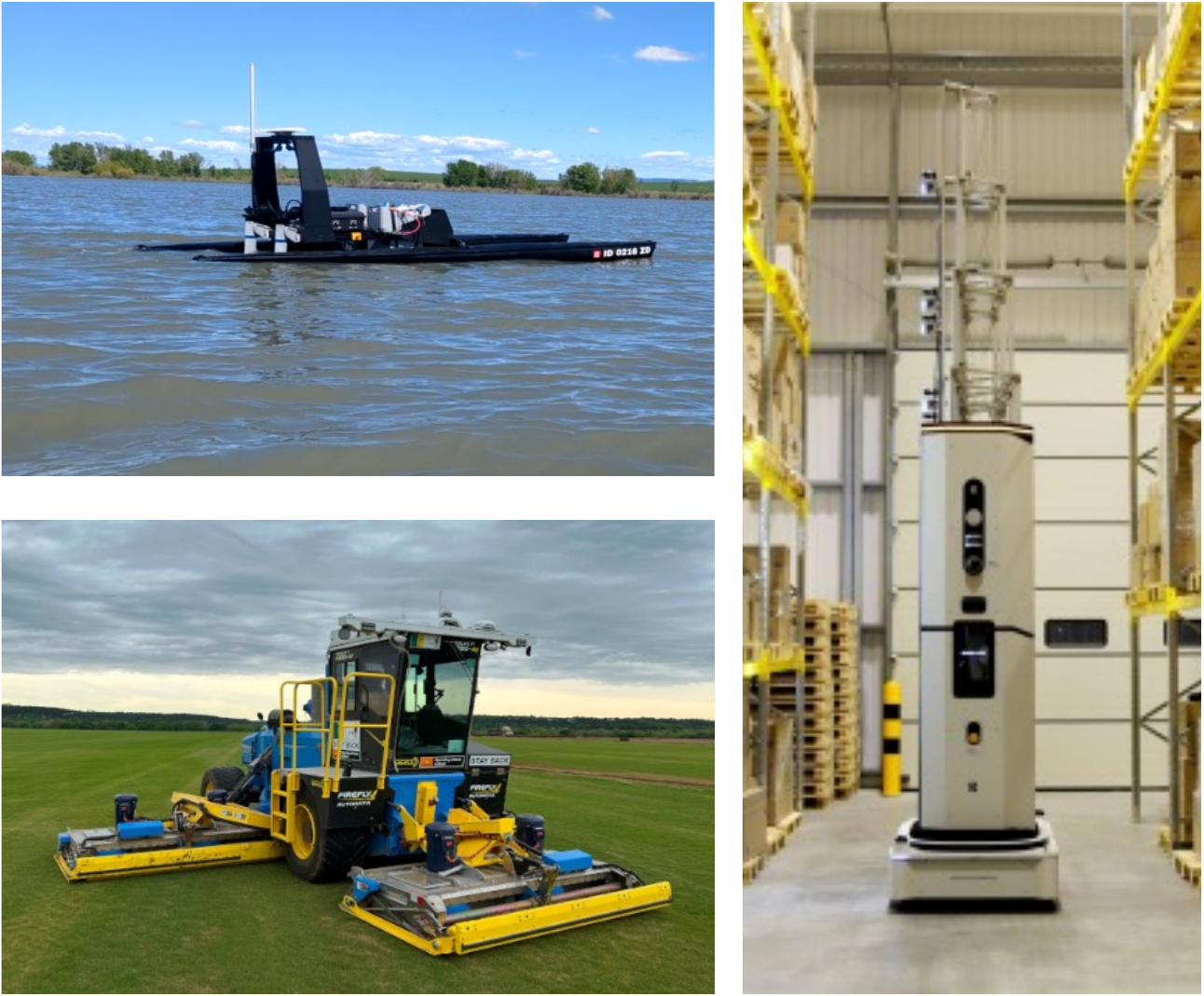}
    \caption{Examples of industrial and marine surface robots using Nav2 uniquely enabled by our \textit{Smac Planner} open-source framework. More examples can be found at \url{https://docs.nav2.org/about/robots.html}.}
    \vspace*{-0.2cm}
    \label{fig:robot}
\end{figure}

We address this critical gap by providing robust planning capabilities for these emerging robot platforms (Fig.~\ref{fig:robot}). Our proposed \textit{Smac Planner} makes three primary contributions:
\begin{enumerate}[leftmargin=\parindent,align=left,labelwidth=\parindent,labelsep=0pt]
    \item \textbf{Developing Cost-Aware planning algorithms} that enhance traditional feasible planners with cost awareness necessary for mobile robotics applications, enabling more efficient navigation in complex and dynamic environments while maintaining kinematic constraints.
    \item \textbf{Integrating these algorithms with widely-used robotics middleware} as the standard planning system within ROS 2 Navigation (Nav2). In use by over 50 organizations worldwide across research, warehouses, outdoor environments, and surface marine applications, \textit{Smac Planner} has enabled ROS to provide support to thousands of Ackermann, legged, humanoid, and large non-circular robots for the first time. %
    \item \textbf{High-performance, open-source implementation} that provides both a template framework for rapidly developing new search-based planners and available implementations of Cost-Aware A*, Hybrid-A*, and State Lattice planners, available at \url{https://github.com/ros-planning/navigation2}. \textit{Smac Planner} is designed to make implementation of search-based planning algorithms simple, requiring only $\approx$200 lines of C++.
\end{enumerate}

\section{Related Work}
\label{sec:related_work}

Kinematically feasible planners model non-circular and/or non-holonomic constraints to provide executable paths for robots with limited maneuverability or arbitrary shapes \cite{kinodynamic}. Two primary search-based approaches dominate this domain: Hybrid-A* and State Lattice planners. Hybrid-A* employs Ackermann curvature-constrained models like Dubins and Reeds-Shepp curves to search grid maps within kinematic limitations \cite{hybrida*}. State Lattice planners apply pre-generated motion primitives (control sets) to find neighbors for arbitrary motion models \cite{lattice, models}.

While sampling-based methods like RRT* exist for kinodynamic planning, they typically demonstrate poor performance over large distances with non-trivial obstacles \cite{rrt*kino}. For this reason, hybrid navigation architectures often restrict kinodynamic planning to local trajectory generation while applying kinematic constraints for global planning \cite{hybridplanning}.

Several planning frameworks have achieved widespread adoption. The Open Motion Planning Library (OMPL)~\cite{ompl} specializes in sampling-based techniques such as Probablistic Roadmaps~\cite{kavraki1996probabilistic} and Rapidly Exploring Random Tree~\cite{lavalle1998rapidly} variants. OMPL exhibits significantly slower run times than search-based methods for low-dimensional, non-trivially occupied spaces and often generates paths with unnecessary turning maneuvers near obstacles. The Search-Based Planning Library (SBPL)~\cite{sbpl} implements heuristic search algorithms like Anytime Repairing A* (ARA*) \cite{ara}. SBPL's manually-engineered motion primitives limit maneuverability, and its naive distance-based heuristics result in slow planning times in structured environments, making it better suited for local trajectory planning rather than global path generation~\cite{sbpl}.

Prior to our work, the primary planner in ROS was Navigation Function (NavFn) \cite{nav2}, which uses cost information in its search with point-cost collision checking. While efficient, NavFn cannot navigate highly asymmetric robots through narrow spaces or provide feasibility guarantees for non-circular robots---a critical gap that \textit{Smac Planner} directly addresses.

\section{Methodology}
\begin{figure}[t]
    \centering
    \includegraphics[width=0.85\linewidth]{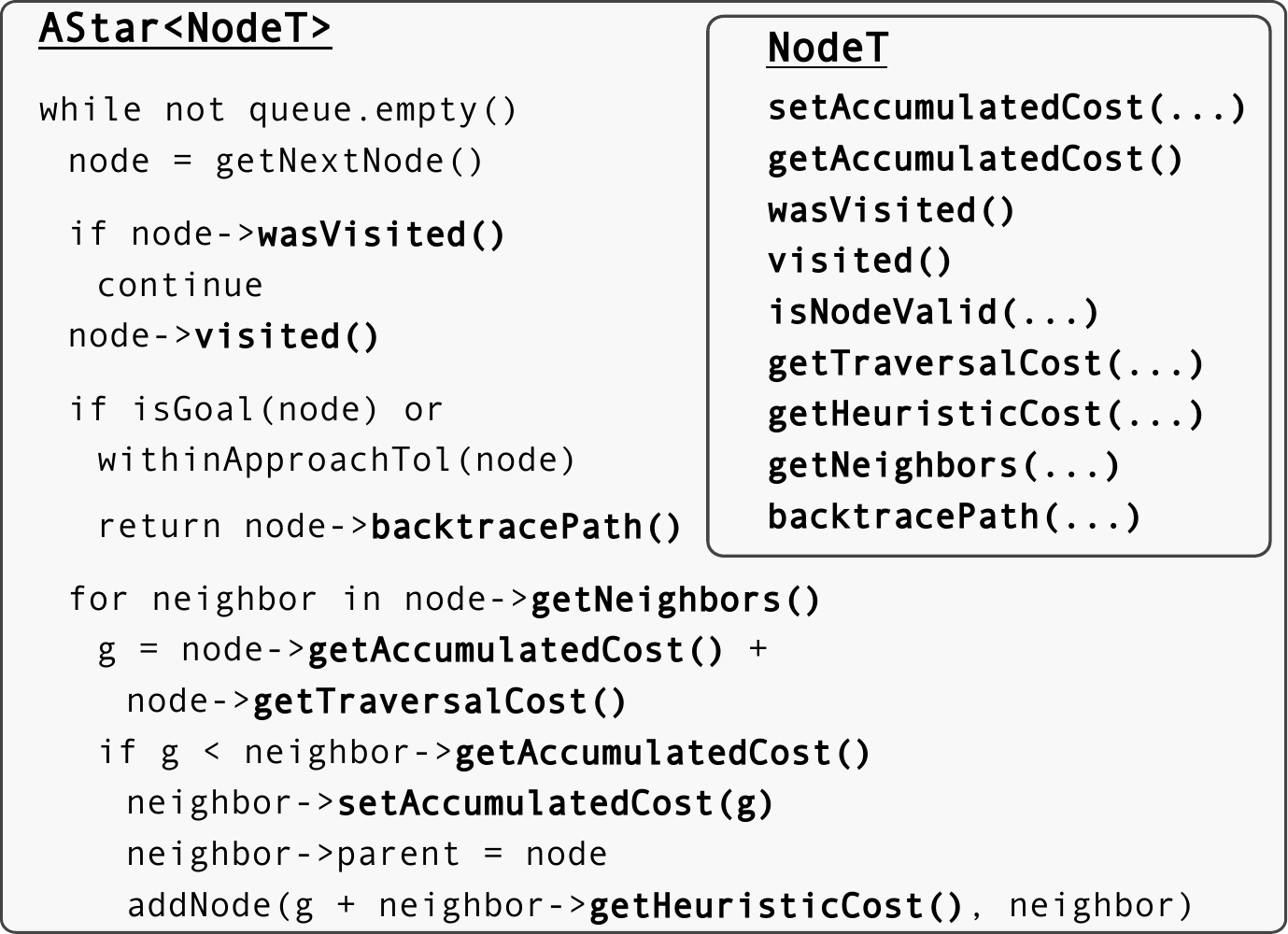}
    \vspace*{-0.15cm}
    \caption{Outline of the A* and key node template methods.}
    \vspace*{-0.4cm}
    \label{fig:design}
\end{figure}

The architecture of Smac Planner consists of three critical components designed to maximize both performance and extensibility. A templated A* is included for search, as it can be easily integrated with arbitrary graph types, search neighborhoods, and heuristic computations required for a variety of planners.
As shown in Fig.~\ref{fig:design}, the A* is templated by the node planner type (\textit{NodeT}) which implements the planner-specific logic. This templated approach enables the creation of varied planners with different characteristics while minimizing boilerplate code. For easy integration with other robotic ecosystems, the \textit{Smac Planner} is isolated from the middleware, with ROS~2 integration occurring only at the highest level, i.e.~the integration layer.

\subsection{Cost-Aware Planning}
\label{sec:mobile}
Our Cost-Aware approach enables mobile robots to navigate efficiently in complex and dynamic environments while maintaining kinematic constraints. 
Mobile robots operate in different contexts than autonomous vehicles for which many feasible planners were originally developed for---frequently with less structured interactions with other agents, fewer compute resources, and differing behavioral constraints.
We formulated the Cost-Aware Obstacle Heuristic to guide feasible planning by incorporating environmental cost information beyond binary obstacle detection (as done in~\cite{hybrida*}). This heuristic performs a 2D-grid search from the goal pose, caching accumulated path costs to guide the kinematically feasible search away from obstacles while respecting soft grid map constraints used to dictate behavioral constraints. It reduces planning time by avoiding dead-ends, U-turns, and inefficient directions using both \textit{obstacle} occupancy and \textit{cost} information from the grid map. It uses a dynamic programming implementation of Differential A* to re-prioritize preexisting queued nodes and restart heuristic search during the planner's execution~\cite{diffA*}.

The traversal cost $C$ incorporates both distance traveled $d$ and a weighted cost-proportional term:
\begin{equation}
  \label{eq:travel_cost}
    C = d \cdot (1 + \frac{\alpha \cdot c_{i,j}}{c_{\max}}),
\end{equation}
where $c_{i,j}$ is the cost of moving from cell $i$ to $j$ and $c_{\max}$ is the maximum value in the cost map. The weight parameter $\alpha$ balances path length with cost avoidance, with higher values directing paths away from high-cost regions at the expense of longer paths. By normalizing the grid cost and applying a weighted penalty, the traversal cost scales proportionally to travel distance. The same heuristic cost in Eq. \ref{eq:travel_cost} is also applied as the path traversal cost in all planners, thereby ensuring that this heuristic is admissible.

\begin{figure}[t]
    \centering
    \includegraphics[trim={0 35cm 0 101cm},clip,width=0.22\textwidth]{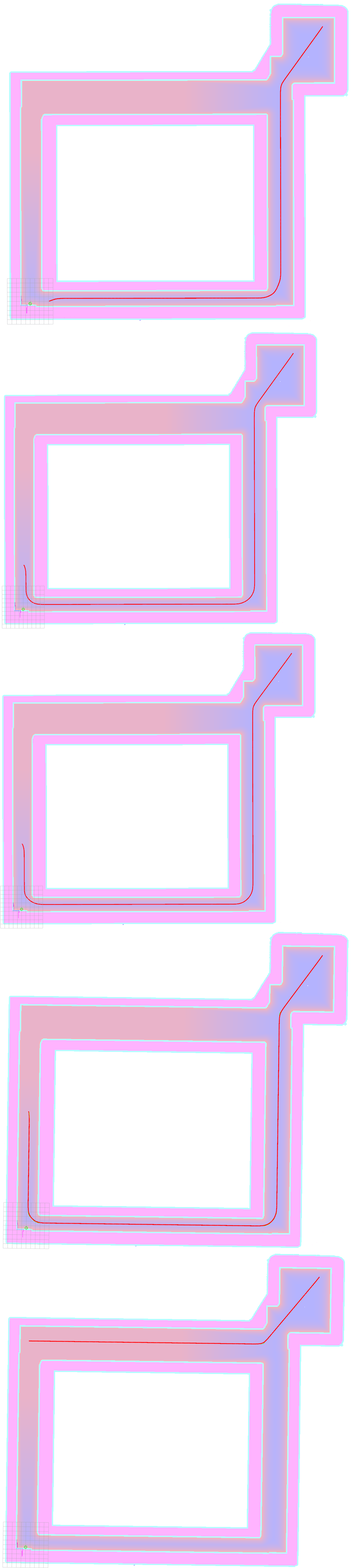}
    \includegraphics[trim={0 0 37.3cm 0},clip,width=0.22\textwidth]{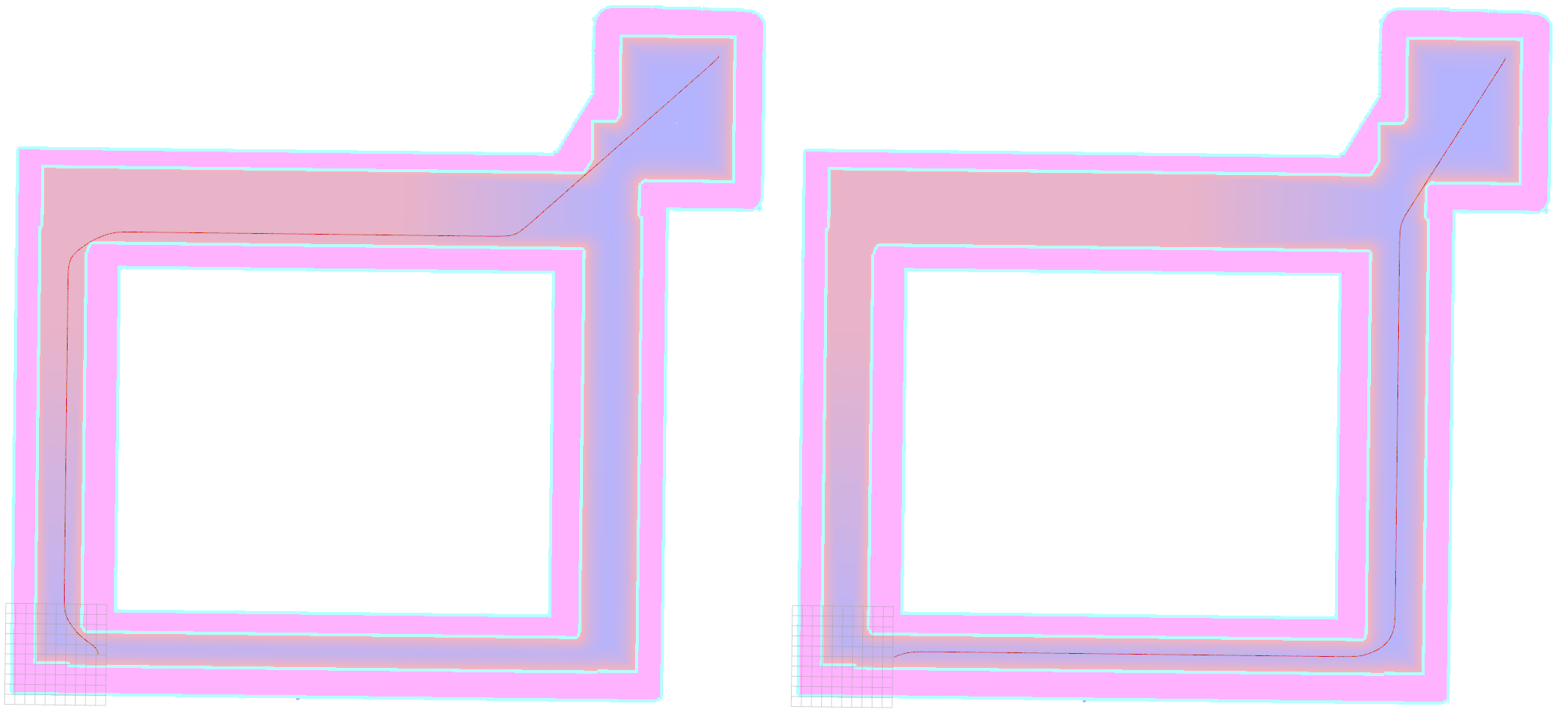}
    \vspace*{-0.2cm}
    \caption{Illustrative example where higher costs are applied to a region to dissuade travel. This soft constraint may be introduced due to increased risk or danger in that area (red).
On the left, the Cost-Aware Obstacle Heuristic uses this constraint and finds a longer path, going around the danger zone to achieve a lower path cost.
On the right, the binary Obstacle Heuristic ignores this constraint and navigates directly through the region for a modestly lower path length.
Our heuristic steered towards the goal and maintained acceptably navigable distances from inflated obstacles.}
\vspace*{-0.3cm}
    \label{fig:cost_region}
\end{figure}

Previous approaches to feasible planning~\cite{hybrida*} typically consider only binary obstacle information and rely on expensive optimization-based path smoothing to maximize obstacle clearance. Our Cost-Aware approach offers significant advantages for mobile robotics, where behavioral constraints are often encoded as cost fields rather than binary obstacles, such as penalizing traversal in front of other dynamic agents or inflating obstacles to punish traversing close to potential collisions. These cost fields can create complex behaviors that prioritize certain regions for exploration.
It is favorable to utilize these during search rather than post-processing to obtain the optimal solution space considering these constraints.
Fig.~\ref{fig:cost_region} shows an illustrative example of a plan that would be difficult to respect the grid-based constraints in a later smoothing stage.
This method also has the added benefit of curtailing the strict requirement for expensive path smoothing.

\subsection{Traversal Penalty Functions}
\label{sec:penalty}
To further enhance path quality, we implement three traversal penalty functions that influence search behavior on top of the cost-aware traversal cost.
The \textbf{non-straight penalty} $\beta$ applies a cost to non-straight motion primitives, reducing unnecessary turning behavior. A small penalty ($0.02\leq\beta\leq0.05$) typically yields smooth, reliable paths. The \textbf{change penalty} $\gamma$ is complementary and punishes changes in turning direction $\delta$ to promote deliberate motion.
These penalties are combined to produce the final traversal cost $C^*$, enabling fine-tuned path behaviors for different mobile robot applications without compromising feasibility:
\begin{equation}
  \label{eq:penalties}
    \begin{aligned}
    C^* =
    &\begin{cases} 
        C & \text{if } \delta_i = 0 \\
        (1 + \beta) \hspace{1mm} C & \text{if } sgn(\delta_i) = sgn(\delta_{i-1}) \\
        (1 + \beta + \gamma) \hspace{1mm} C &  \text{if }sgn(\delta_i) \neq sgn(\delta_{i-1})
    \end{cases}
    \end{aligned}
\end{equation}

Finally, the \textbf{reverse penalty} discourages motion primitives in the reverse direction, allowing backing up only when necessary for maneuvers like turning around. 

\subsection{Search Resolution}
\label{sec:resolution}
Early Hybrid-A* used coarse search grids (0.5m) on finer obstacle maps (0.15m), effectively increasing primitive lengths by $\sim$3.3x. Hybrid-A* then relied on expensive smoothing to recover acceptable paths, which creates unstable search behavior and complicates planning in narrow passages. Smaller primitives at grid resolution create more reliable paths without requiring extensive smoothing. \textit{Smac Planner} achieves comparable compute performance while maintaining the obstacle grid resolution for search in all planners.%

\subsection{Path Planners}
\label{sec:implementation}
Three planners are implemented within the framework: 2D-A*, Hybrid-A*, and State Lattice. Each is built as node planner types on the templated A* algorithm and is implemented in under 300 lines of planner-specific code, demonstrating the efficiency of our approach.

\subsubsection{Cost-Aware 2D-A* Planner}
\label{sec:2d}
The Cost-Aware 2D-A* provides baseline functionality for circular robots. It employs the same traversal and heuristic properties as the Cost-Aware Obstacle Heuristic (Section~\ref{sec:mobile}) but is structured as an end-user planner. The planner uses Moore (8) connected spaces for its search with a distance heuristic. Suppl.~Fig.~\ref{fig:2d} shows the Cost-Aware 2D-A* in a warehouse environment finding paths approximately 85m in length.

\subsubsection{Cost-Aware Hybrid-A* Planner}
\label{sec:hybrid}
The Cost-Aware Hybrid-A* planner provides feasible paths using Dubins or Reeds-Shepp motion models constrained by a minimum turning radius for Ackermann vehicles. It stores continuous coordinates of expanded nodes to create drivable paths. This planner typically executes in 20-300ms across warehouse environments of approximately 6500$m^2$, achieving speeds as fast as 5ms when the heuristic is cached for replanning.
The planner uses two admissible heuristics, of which it takes the maximum: the Cost-Aware Obstacle heuristic and a non-holonomic-without-obstacles heuristic from~\cite{hybrida*} using a pre-computed lookup table. The latter heuristic prunes approach headings that would be difficult to achieve from the drivetrain's constraints. %

Motion primitives are dynamically computed proportional to grid map resolution, with the minimum required angular change determined by the angle of a chord at least $\sqrt{2}$ cells long from a circle with the minimum turning radius. Our implementation uses diverse primitives beyond simple left, right, and straight motions to fully populate the quantization search space, which improves runtime efficiency, performs better in constrained environments, and generates higher quality paths. We support multiple goal orientation options: respecting exact goal orientation, bidirectional orientation (goal, goal+180°) for orientation-agnostic arrival of bidirectional systems, and non-orientation respecting paths that minimize total path length. For the latter, we employ a coarse-to-fine search in the analytic expansion module to find the most efficient valid path. Analytic expansion allows the search process to continue far from the goal to account for grid map behavior constraints.

\subsubsection{Cost-Aware State Lattice Planner}
\label{sec:lattice}
The Cost-Aware State Lattice planner is based upon~\cite{minimalset,lattice} and searches using pre-generated motion primitives for arbitrary motion models (e.g.~Ackermann, differential, omnidirectional). Its computational performance is comparable to Hybrid-A* at 10-350ms in similar environments, improving to 1-3ms with cached heuristics. Our implementation generates principled minimum control sets that represent the state lattice of a motion model with the smallest possible set of motions using a minimum turning radius constraint.

The state lattice planner uses the same admissible heuristics, pre-computation optimizations, and traversal functions as the Cost-Aware Hybrid-A*. The analytic expansion and Cost-Aware heuristic are new to the state lattice planner regime and significantly decrease planning times.
Rather than employing complex path smoothing that might invalidate cost constraints, we implement a lightweight gradient descent approach that balances smoothness with preservation of the original path's intent. This smoother typically completes in 1-6ms.%

\begin{figure}[t]
    \centering
    \includegraphics[width=0.6\linewidth]{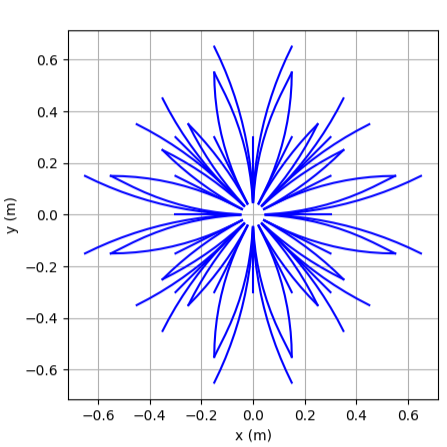}
    \vspace*{-0.2cm}
    \caption{A control set using a 5cm grid resolution and a 1m turning radius with 16 angle quantizations. Each  quantization angle has 3-5 unique primitives.}
\vspace*{-0.3cm}
    \label{fig:controlset}
\end{figure}

The Supplementary Material details the minimal control set generation, which is the smallest set of motions needed to represent the state lattice of a motion model (see Fig.~\ref{fig:controlset}). %

\section{Experiments and Analysis}
\label{sec:experiments}

We evaluated the \textit{Smac Planner} in both controlled random environments and a real-world warehouse setting, comparing performance against established baselines. 

\subsection{Simulation Experiments}
\label{sec:sandbox}
We constructed a comparative benchmark of the three \textit{Smac Planner} implementations against established baselines, namely NavFn~\cite{nav2} and SBPL's ARA*~\cite{sbpl}, across three 10,000$m^2$ random environments with varying obstacle densities (10\%, 15\%, 20\%). For each environment, we generated 1,000 verified start-goal pairs with minimum separation of 3m. All experiments used an AMD Ryzen 5 5600X CPU. The planners were configured with consistent parameters: minimum turning radius of 0.4m, reversing enabled, cost penalty $\alpha=2.0$, and non-straight and change penalties $\beta=\gamma=0.05$. SBPL's parameters were equivalently set with 16 heading discretizations, 5cm grid resolution, and 5 primitives per heading.

Table \ref{tab:occ} shows that the feasible \textit{Smac Planners} (Hybrid-A* and State Lattice) demonstrated consistently superior performance across all obstacle densities. Both feasible planners exhibited remarkable consistency, with execution times between 39-42ms and less than 5\% variation in runtime across all scenarios. They outperformed Cost-Aware 2D-A* by $\approx$50\% and NavFn by 38\% in terms of execution time. While Cost-Aware 2D-A* produced the shortest paths, the feasible planners generated paths within 2.5\% of optimal length -- a modest increase attributable to kinematic constraints. SBPL's implementation required approximately two orders of magnitude greater computation time than the Cost-Aware State Lattice counterpart, primarily due to inefficient goal approach behavior. SBPL also has longer paths than any of the proposed planners that regularly have more sharp, unsmooth turning behaviors.

Suppl.~Fig.~\ref{fig:example_paths} illustrates characteristic paths from each planner. The 2D-A* produces mechanical paths with straight-line segments, while Hybrid-A* and State Lattice generate smooth, feasible paths with appropriate obstacle clearance. The NavFn planner creates relatively smooth paths but lacks feasibility guarantees and exhibits slower performance.

\begin{table}[t]
\caption{Random occupancy map results. Best results and those within 1\% (relative) are marked in bold.}
\vspace*{-0.2cm}
\centering
\footnotesize
\setlength{\tabcolsep}{3pt}
\begin{tabular}{c|c|ccccc} 
\hline
\multirow{2}{*}{Obstacles} & \multirow{2}{*}{Metric} & \multicolumn{5}{c}{Planner} \\
 & & Hybrid-A* & State Lattice & 2D-A* & SBPL & NavFn \\
\hline
\multirow{2}{*}{$10\%$} & $t$ (ms) & \textbf{39.07} & $42.31$ & $66.15$ & $5,640$ & $71.11$ \\ 
& $l_{path}$ (m) & \textbf{51.41} & \textbf{51.40} & \textbf{50.96} & $51.99$ & $52.60$ \\ 
\hline
\multirow{2}{*}{$15\%$} & $t$ (ms) & \textbf{40.73} & $43.25$ & $85.63$ & $6,587$ & $66.45$\\ 
& $l_{path}$ (m) & $51.10$ & $51.15$ & \textbf{50.45} & $54.87$ & $52.50$\\ 
\hline
\multirow{2}{*}{$20\%$} & $t$ (ms) & \textbf{38.77} & $39.40$ & $88.82$ & $6,633$ & $61.02$\\ 
& $l_{path}$ (m) & $50.78$ & $50.51$ & \textbf{49.65} & $53.68$ & $52.25$\\
\hline
\end{tabular}
\vspace*{-0.25cm}
\label{tab:occ}
\end{table}

\subsection{Real-World Experiments}
\label{sec:warehouse}

To validate performance in a challenging real-world setting, we deployed the \textit{Smac Planners} in a 33,600$m^2$ (363,000$ft^2$) warehouse environment (Suppl.~Fig.~\ref{fig:locus}), using the same parameters as in Sec.~\ref{sec:sandbox} but using an Intel i7-8565U CPU. %
Three goal poses were selected to evaluate performance in both major thoroughfares and confined aisles.

Feasible planners demonstrated superior qualitative performance in this complex environment: 2D-A* paths exhibited unnatural movements outside main thoroughfares, while Hybrid-A* and State Lattice produced high-quality paths even in challenging confined areas and smoothly transitioning between aisles. Quantitative results averaged over 10 planning trials confirmed the computational advantage of the feasible planners, with Hybrid-A* requiring 290ms, State Lattice 473ms, and 2D-A* 1358ms. This performance hierarchy aligns with our controlled experiments but reflects the increased computational demands of the vastly larger and more complex environment.

The experimental results confirm our hypothesis that Cost-Aware planning significantly reduces computational burden while maintaining near-optimal path quality in both controlled and real-world settings. The comparable behavior between Hybrid-A* and State Lattice suggests that selection between them can be based primarily on the desired motion model rather than performance considerations.

\section{Conclusion}
\label{sec:conclusion}
This paper presented \textit{Smac Planner}, a high-performance search-based planning framework with Cost-Aware variations specifically designed for mobile and surface robotics applications. Our key innovation---incorporating cost awareness throughout the planning process---enables more effective navigation for non-circular, Ackermann, and legged robots. Experimental results demonstrated that our feasible planners consistently outperform baseline methods, with Hybrid-A* and State Lattice implementations executing faster with additional advantages in complex modern applications.

The \textit{Smac Planner} addresses a gap in robotics by providing kinematically feasible planners integrated with mainstream frameworks. Its adoption as the standard planning system in Nav2 has enabled previously unsupported robot classes to navigate complex environments, with documented deployments across delivery, warehouse, and marine applications worldwide. The implementation is freely available, providing the robotics community with practical tools for deploying sophisticated autonomous navigation capabilities across diverse platforms.

\renewcommand{\figurename}{Suppl.~Fig.}
\setcounter{figure}{0}
\begin{figure*}[htb]
    \centering
    \includegraphics[width=0.78\textwidth]{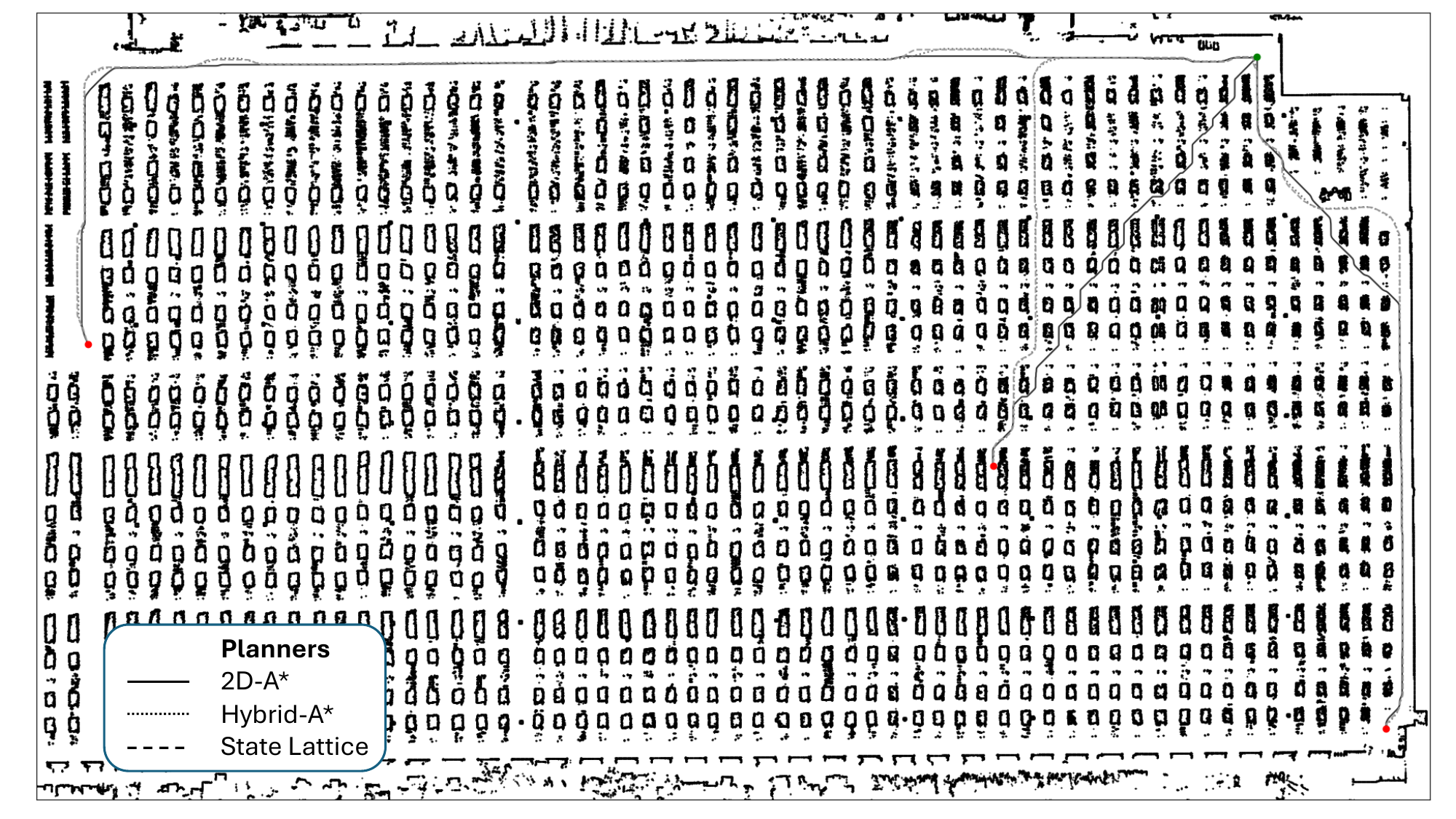}
    \caption{A set of example paths of the \textit{Smac Planner} in a 33,600$m^2$ warehouse serviced by Locus Robotics.}
    \label{fig:locus}
\end{figure*}
\section*{Supplementary Material}
\paragraph*{Code and documentation}
The code for our \textit{Smac Planner} is hosted in the \textit{nav2\_smac\_planner} subdirectory of the \url{https://github.com/ros-navigation/navigation2} repository. %
The \textit{README.md} file contains important side notes, including %
setting the turning radius for the Hybrid-A* and State Lattice planners, %
and properly tuning the penalty functions; including a link to a configuration guide that contains descriptions of additional parameters. %

\paragraph*{Reproducing results}
Results for the simulation experiments can be replicated using our planning benchmark suite at \url{https://github.com/ros-navigation/navigation2/tree/main/tools/planner_benchmarking}.

\paragraph*{Talk recording}
A recording at ROSCon covers key points of the planner, demonstrations, and how to enable the planner in an application, which can be accessed at \url{https://vimeo.com/showcase/9954564/video/767157646}.

\paragraph*{Minimal control set generation}
Minimal control sets are generated by evaluating feasible trajectories from the origin to end poses across different start headings. Trajectories are added only if they cannot be decomposed by existing paths, with shorter paths generated first to facilitate decomposition of subsequent paths. End points are selected based on a wavefront expanding outward from the origin. The process terminates when all trajectories in several consecutive wavefronts become decomposable. Rather than using uniform heading discretization, which produces excessively long non-straight paths \cite{minimalset}, we derive non-uniform heading angles from cell centers in the wavefront. This approach creates shorter, more efficient motions that decompose into fewer trajectories, resulting in a more practical control set for representing the robot's motion capabilities.

\begin{figure}[ht] 
    \centering
    \includegraphics[width=0.9\linewidth]{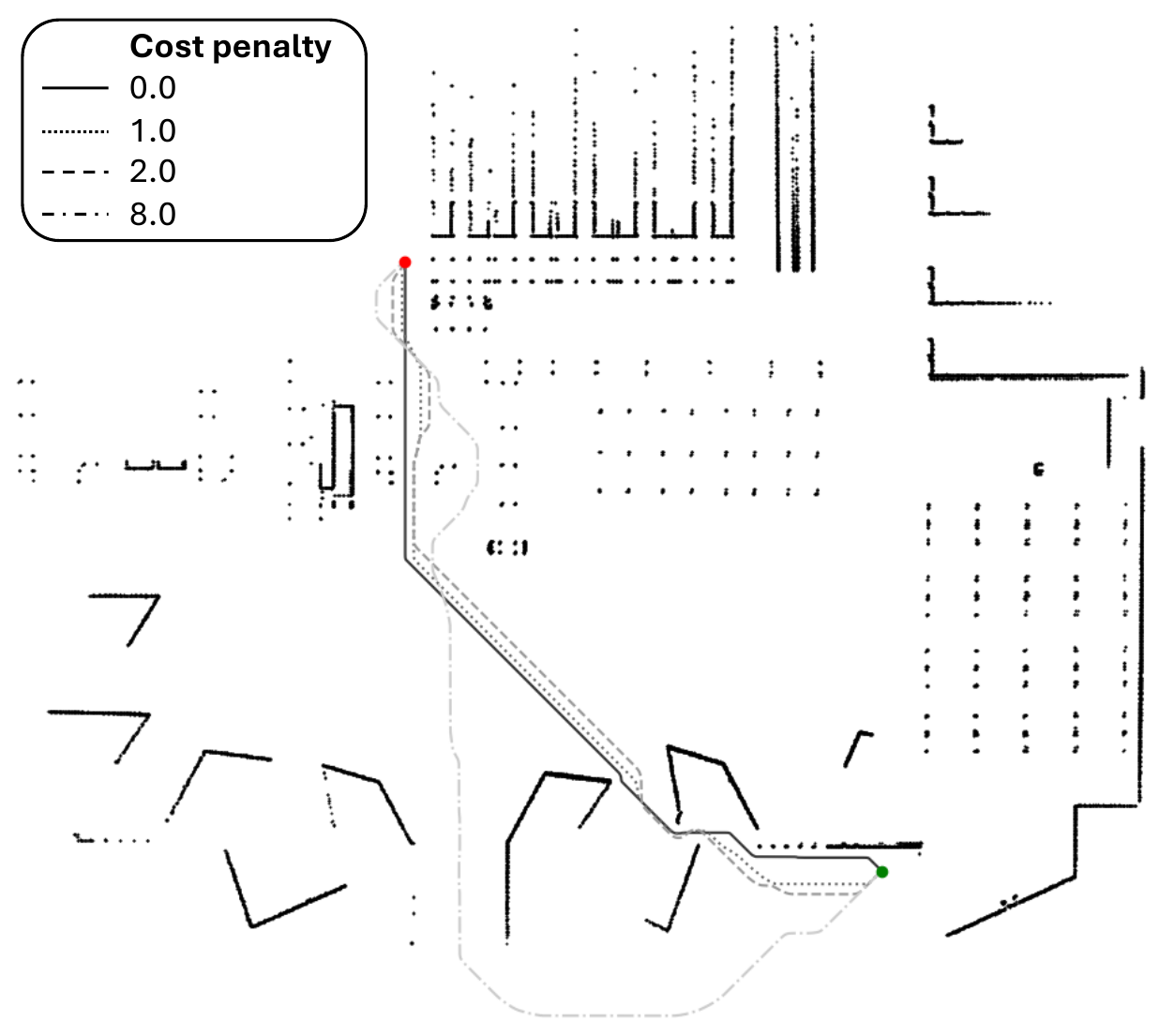}
    \vspace*{-0.3cm}
    \caption{Cost-Aware 2D-A* with variable cost penalties.}
    \label{fig:2d}
\end{figure}

\begin{figure}[ht]
    \centering
    \includegraphics[trim={12cm 8.5cm 12cm 0cm},clip,width=0.9\linewidth]{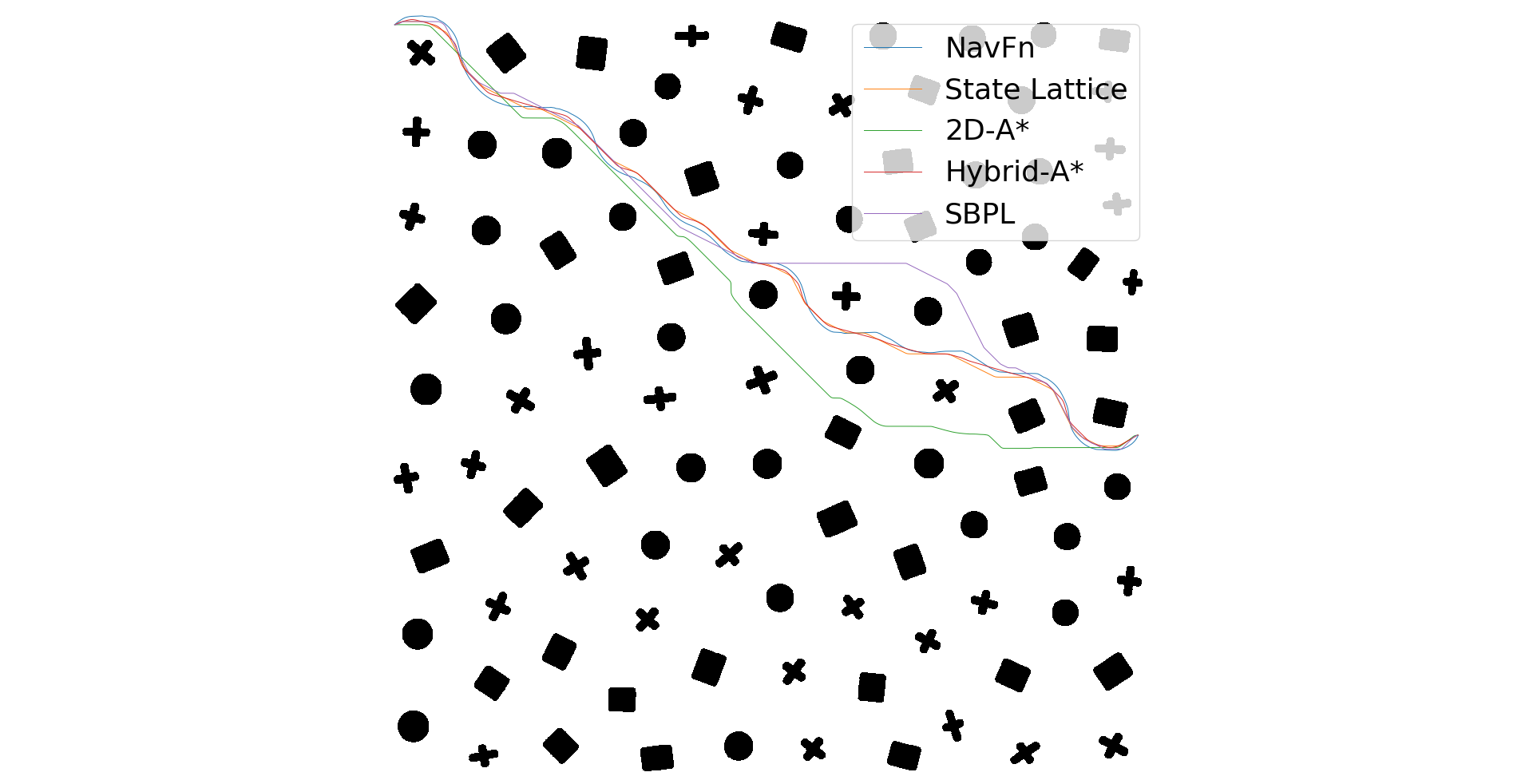}
    \caption{Example paths in the 20\% occupied map.}
    \label{fig:example_paths}
\end{figure}

\ifarxiv
\paragraph*{Generating trajectories in the set}
To generate control sets, we must generate candidate feasible trajectories to any given state. 
We propose an intuitive method for trajectory generation that guarantees minimal curvature (Suppl.~Fig.~\ref{fig:trajgen-explanation}).
Given a start configuration $(x_I,y_I,\theta_I)$ and an end configuration $(x_F, y_F, \theta_F)$, we define two lines, $l_1$ which passes through the start configuration and $l_2$ which passes through the end configuration:
\begin{align}
    l_1(x) &= (x - x_I)\,\tan\theta_I + y_I \\
    l_2(x) &= (x - x_F)\tan\theta_F + y_F
\end{align}

\begin{figure}[htb]
    \centering
    \includegraphics[width=0.8\linewidth]{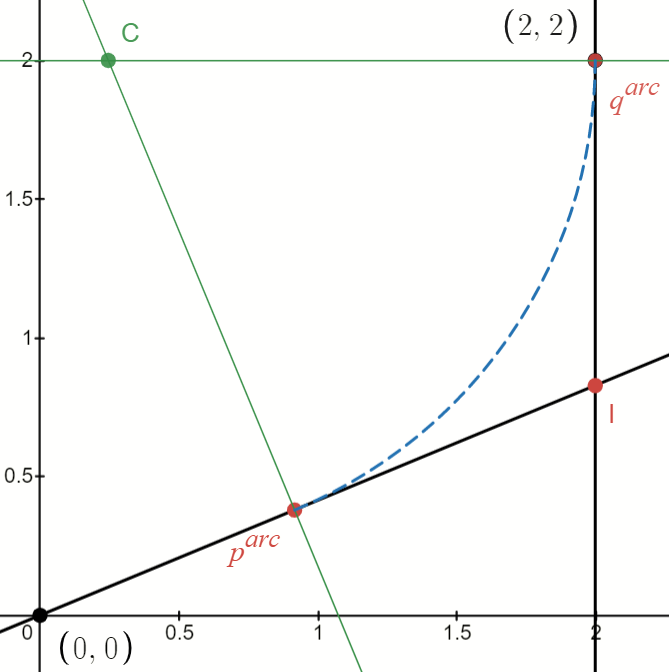}
    \caption{An example trajectory for start configuration (0,0,$\frac{\pi}{8}$) to end configuration (2,2,$\frac{\pi}{2}$).}
    \label{fig:trajgen-explanation}
\end{figure}

Then, we compute their intersection point $I$:
\begin{align}\label{eq:intersection_point}
    a &= \frac{y_F - x_F \tan\theta_F - y_I + x_I\tan\theta_I}{\tan\theta_I - \tan\theta_F}\\
    I &= [a, l_1(a)]
\end{align}

Next, we calculate the end points of the trajectories arc. $p^{arc}$ will lie on $l_1$ and $q^{arc}$ will lie on $l_2$. The end points are calculated using distance $d$, the minimum of either $I$ to $p$ or $I$ to $q$, where $p = (x_I, y_I)$ and $q = (x_F, y_F)$.
By nature of $d$, either $p^{arc}$ will be coincident with $p$, or $q^{arc}$ will be coincident with $q$:

\begin{align}\label{eq:arc_endpoint}
    d &= \min(\lVert I - p \rVert, \lVert I - q \rVert) \\
    p^{arc} &= d \: \overrightarrow{p-I} + I \\
    q^{arc} &= d \: \overrightarrow{q-I} + I
\end{align}

Finally, we find the center point $C$ and radius $r$ of the circle whose arc joins $p^{arc}$ at an angle of $\theta_I$, to $q^{arc}$ at $\theta_F$.
This is found via an intersection point of perpendicular bisectors of $l_1$ at $p^{arc}$ and $l_2$ at $q^{arc}$.
This is the center since the distances from $p^{arc}$ and $q^{arc}$ are $r$: 

\begin{align}\label{eq:circle_params}
    l_1^\perp(x) &= x\frac{-1}{\tan\theta_I} + p^{arc}_y + \frac{p^{arc}_x}{\tan\theta_I} \\
    l_2^\perp(x) &= x\frac{-1}{\tan\theta_F} + q^{arc}_y + \frac{q^{arc}_x}{\tan\theta_F} \\
    b &= \frac{q^{arc}_y + \frac{q^{arc}_x}{\tan\theta_F} - p^{arc}_y - \frac{p^{arc}_x}{\tan\theta_I}}{\frac{-1}{\tan\theta_I} + \frac{1}{\tan\theta_F}} \\
    C &= [b, l_1^\perp(b)]\\
    r &= \lVert C - p^{arc} \rVert
\end{align}

To complete the trajectory, a straight-line segment is added connecting either the start or end configuration to the arc.
Using this method, the trajectory is either an arc or contains an arc and a line segment, depending on the inputs, which guarantees that the trajectory contains the minimal curvature.

An additional step ensures that the minimal curvature is within the vehicle's physical constraint, the minimum turning radius $R$. 
The closer $q^{arc}$ and $p^{arc}$ are to $I$, the smaller the radius of the path.
This places a lower limit on $d$ for the trajectory to be valid.
The minimum distance $d_{\min}$ is then:
\begin{align}\label{eq:radius_constraint}
    d_{\min} = & \frac{R}{\tan\frac{\psi}{2}}, \text{where}\\
    \psi = & \frac{\pi}{2} - |\theta_F - \theta_I|.
\end{align}

\IEEEtriggeratref{18}
\else
\clearpage
\raggedbottom
\IEEEtriggeratref{13}
\fi

\bibliographystyle{IEEEtran}
\bibliography{references}

\end{document}